\documentclass[11pt]{article}

\usepackage{acl}

\usepackage{times}
\usepackage{latexsym}

\usepackage[T1]{fontenc}
\usepackage[utf8]{inputenc}

\usepackage{microtype}
\usepackage{amsmath}
\usepackage{amssymb}
\usepackage{booktabs}
\usepackage{graphicx}
\usepackage{tabularx}
\usepackage{algorithm}
\usepackage{algorithmic}
\newcolumntype{Y}{>{\raggedright\arraybackslash}X}
\newcommand{\Sel}{\operatorname{Sel}}
\newcommand{\Stair}{\operatorname{Stair}}
\newcommand{\RetAvg}{\operatorname{RetAvg}}
\newcommand{\HardAvg}{\operatorname{HardAvg}}
\newcommand{\Leak}{\operatorname{Leak}}
\newcommand{\Emerg}{\operatorname{Emerg}}
\newcommand{\RecallOne}{\operatorname{R@1}}

\usepackage{inconsolata}

\title{GraSP-VL: Length as a Semantic Granularity Interface for Vision-Language Representations}

\author{
Zesheng Li, Chengchang Pan\thanks{Co-corresponding authors.}, and Honggang Qi\footnotemark[1] \\
University of Chinese Academy of Sciences, Beijing, China \\
\texttt{2022111515@stu.sufe.edu.cn} \\
\texttt{166353314@qq.com; hgqi@ucas.ac.cn}
}

\begin{document}
\maketitle

\begin{abstract}
Frozen vision-language models already encode object, attribute, relation, and caption information, but a single vector exposes all of these signals through the same access pattern. We ask whether vector length can become a semantic access control: short prefixes answer coarse queries, while longer prefixes add finer distinctions. GraSP-VL learns one shared near-orthogonal transform for image and text embeddings and assigns a fixed, configurable prefix contract. At inference, the transform is applied once and ordinary cosine retrieval uses the prefix required by the task. On a 20,147-example COCO/Flickr30K pool, GraSP-VL reaches a staircase score of 53.01 and hard-negative selectivity of 89.76 with full-space drift below $10^{-6}$. Full-dimensional ImageNet-V2 classification and Flickr30K/COCO Karpathy retrieval are unchanged, while held-out human-caption retrieval improves at shorter prefixes. These results show that length can reorganize access to frozen VLM semantics rather than merely compressing them. Code and project page are available at \url{https://lizesheng13.github.io/Grasp-VL/}.
\end{abstract}

\section{Introduction}

Vision-language models (VLMs) such as CLIP make one embedding space support caption retrieval, zero-shot recognition, and prompts at very different semantic resolutions~\citep{radford2021learning,jia2021scaling,zhai2022lit,cherti2023reproducible}.
    This is convenient, but it hides an interface mismatch: sometimes a query only asks for an entity such as \textit{dog}; sometimes it asks for an attributed phrase such as \textit{brown dog}; sometimes the crucial information is a relation or event such as \textit{dog running through snow}.
A monolithic embedding gives all of these questions the same access pattern.
Our premise is that frozen VLM outputs contain multi-granularity semantic signals; the missing piece is a controllable way to access them.

\begin{figure}[t]
\centering
\includegraphics[width=\columnwidth]{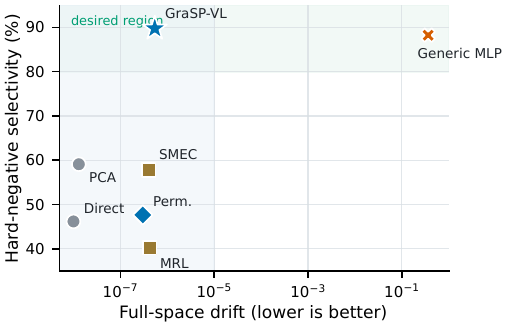}
\caption{Teaser: selectivity--preservation trade-off. GraSP-VL reaches the desired upper-left region, combining high hard-negative selectivity with near-zero full-space drift.}
\label{fig:teaser}
\end{figure}

The usual way to make embeddings shorter is to truncate, project, or compress them.
These operations may preserve performance, but they do not tell us what the shorter vector is supposed to mean.
The first small fraction of a frozen CLIP embedding is not guaranteed to be object-level; PCA may keep strong variance directions, but it does not know that a prefix should represent entities before attributes or relations.
In this paper we study a different interface question: can representation length itself indicate semantic granularity?

This question is close to, but distinct from, Matryoshka-style representation learning~\citep{kusupati2022matryoshka,yoon2024matryoshka,zhang2025smec}, which makes multiple prefix lengths useful under different storage or latency budgets.
Our target is purposeful semantics under truncation: a short prefix should be coarse, and a longer prefix should add distinctions that the shorter one omits.
Put differently, length should be a semantic lens, not only a compression budget.

We propose \textbf{GraSP-VL} (\textbf{Gra}nularity-\textbf{S}elective \textbf{P}refixes for \textbf{V}ision-\textbf{L}anguage representations), a method for learning this access interface over frozen VLM embeddings.
We use \textbf{Semantic Matryoshka} to denote the resulting prefix-interface principle: shorter prefixes are intentionally coarse, while longer prefixes reveal progressively finer distinctions under an explicit contract.
GraSP-VL keeps the VLM frozen and learns a lightweight transform that is shared by image and text embeddings.
The key constraint is that the full-dimensional geometry is preserved, while prefix geometry changes.
We test whether a shared transform can reorganize information already present in the frozen embedding so that short prefixes match coarse language and longer prefixes add increasingly compositional distinctions.
The dimensional breakpoints are configurable interface settings, and the original coordinate order is not assumed to encode a semantic hierarchy.

Our evaluation separates four properties. We first measure multi-granularity signal in frozen VLM embeddings. We then compare raw coordinates, PCA, rotations, compression objectives, and adapters. We test whether GraSP-VL realizes the assigned prefix staircase and whether it transfers across prompts, captions, and backbones. This design separates semantic organization from generic compression or adaptation.

Our contributions are:
\begin{itemize}
    \item We propose \textbf{GraSP-VL}, a shared prefix transform for frozen VLM embeddings that preserves full-space similarity while changing which semantics are exposed by short prefixes.
    \item We formulate \textbf{Semantic Matryoshka} as the access-interface principle implemented by GraSP-VL: embedding length is assigned semantic granularity by an explicit contract rather than treated as a discovered natural boundary.
    \item We introduce a granularity-selective objective that combines prefix capability, cumulative retention, hard-negative discrimination, early-prefix invariance, and full-space preservation.
    \item We design a diagnostic protocol for prefix organization that separates semantic selectivity from raw prefix truncation, PCA compression, Matryoshka-style preservation, generic adaptation, and single-embedding compositional tuning.
\end{itemize}

\begin{figure*}[t]
\centering
\includegraphics[width=\textwidth]{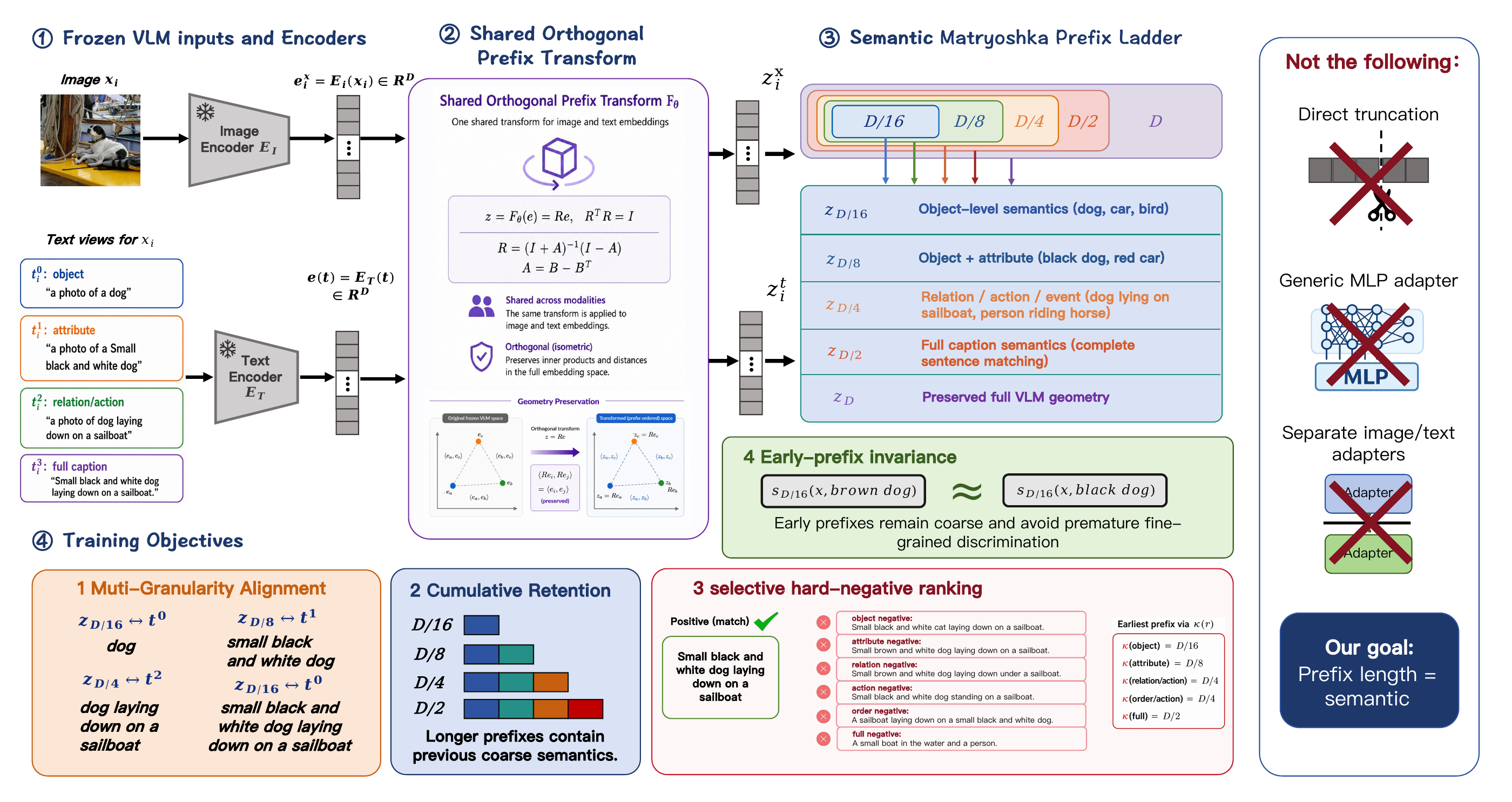}
\caption{Overview of GraSP-VL. Frozen image and text embeddings share one orthogonal transform $R$. Nested prefixes of the transformed vectors are assigned object, attribute, relation/action, and full-caption responsibilities, while the full vector preserves the original VLM geometry.}
\label{fig:framework}
\end{figure*}

\begin{figure*}[t]
\centering
\includegraphics[width=0.9\textwidth]{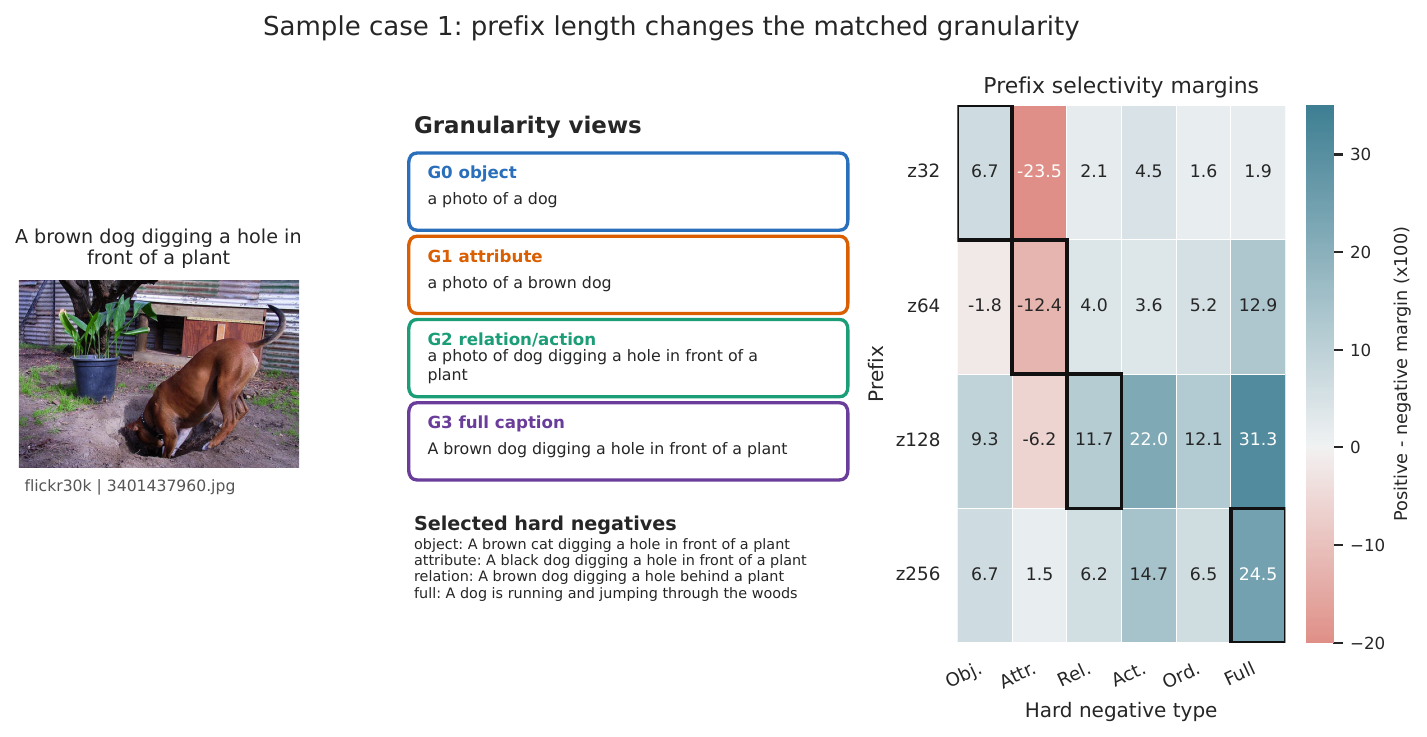}
\caption{Sample case. Longer prefixes progressively expose attribute, relation/action, and full-caption distinctions.}
\label{fig:sample-case}
\end{figure*}

\section{Related Work}

\paragraph{Nested representations.}

Nested representations make one embedding usable at multiple dimensionalities.
Matryoshka Representation Learning trains prefixes of the same vector to remain predictive~\citep{kusupati2022matryoshka}, and Matryoshka-Adaptor adapts pretrained embeddings for smaller usable dimensions~\citep{yoon2024matryoshka}.
Retrieval-specific work makes the compression motivation explicit: MRL-AdANNS uses Matryoshka representations for adaptive web-scale semantic search~\citep{rege2023mrladanns}, and SMEC re-evaluates MRL for retrieval embedding compression with sequential training, adaptive dimension selection, and cross-batch memory~\citep{zhang2025smec}.
Recent multimodal nested systems connect this idea to token-set or multi-vector retrieval~\citep{li2024twodmatryoshka,cai2024m3,xiao2025metaembed}.
These methods target prefix usability and adaptive retrieval budgets; GraSP-VL instead assigns different semantic responsibilities to different prefixes.

\paragraph{VLM adaptation.}

Many methods adapt frozen or partially frozen VLMs with a small number of trainable parameters.
Prompt-learning methods such as CoOp, CoCoOp, and MaPLe tune textual or multi-modal prompts~\citep{zhou2022coop,zhou2022cocoop,khattak2023maple}.
Feature-side methods such as CLIP-Adapter and Tip-Adapter modify or cache CLIP features for downstream recognition~\citep{gao2021clipadapter,zhang2022tipadapter}, while adapters and low-rank adaptation provide general parameter-efficient machinery~\citep{houlsby2019parameter,hu2022lora}.
GraSP-VL is also lightweight, but it reorganizes prefix access while preserving full-space behavior rather than adapting the VLM to a downstream classifier or domain.

\paragraph{Compositional VLMs.}

Standard retrieval scores often hide whether a VLM understands attributes, relations, or word order.
SugarCrepe probes object, attribute, relation, and hard caption distractors outside standard retrieval leaderboards~\citep{hsieh2023sugarcrepe}.
Recent tuning methods such as NegCLIP, CE-CLIP, FSC-CLIP, CLIP-CAE, TripletCLIP, CLoVe, AHNPL, and DeGLA improve compositional sensitivity through hard negatives, preservation-aware training, attribution enhancement, model patching, or global-local alignment~\citep{yuksekgonul2023when,zhang2024ceclip,oh2024preserving,li2024clipcae,patel2024tripletclip,castro2024clove,huang2025ahnpl,hu2025degla}.
Fine-grained retrieval models such as SCAN and FILIP use region-word or token-level interactions rather than a single truncatable vector~\citep{lee2018scan,yao2022filip}; FG-CLIP and FineLIP strengthen CLIP-style full embeddings with fine-grained data~\citep{xie2025fgclip,asokan2025finelip}.
These methods improve compositionality in one representation, while we ask where such sensitivity should appear across prefixes.

\paragraph{Hierarchical semantics.}

Hierarchical semantics have long been central to language and vision.
WordNet organizes lexical concepts taxonomically~\citep{miller1995wordnet}, while ImageNet and Visual Genome provide visual categories and dense semantic relations~\citep{deng2009imagenet,krishna2017visualgenome}.
Representation learning has also used order embeddings, Poincar\'e embeddings, and hyperbolic image-text spaces to model hierarchy and asymmetric relations~\citep{vendrov2016order,nickel2017poincare,desai2023meru}.
Recent VLM analyses further show that CLIP-style embeddings encode fine-grained but uneven object and order information~\citep{abbasi2025clipmicroscope}, while Insight extracts interpretable, spatially grounded semantic hierarchies from vision-language encoders using hierarchical concept representations~\citep{wittenmayer2026insight}.
GraSP-VL is complementary: it does not extract a taxonomy, but learns a Semantic Matryoshka prefix interface in a frozen Euclidean VLM space.

\section{Method}

\subsection{Problem Setup}

Let $E_I$ and $E_T$ be frozen image and text encoders from a pretrained VLM.
For an image $x_i$ and a text string $t$, we denote their normalized embeddings as
\[
\begin{aligned}
e_i^x &= E_I(x_i),\\
e(t) &= E_T(t),\\
e_i^x,e(t) &\in\mathbb{R}^D .
\end{aligned}
\]
Each training pair has language views $t_i^0$--$t_i^3$ for object, object+attribute, relation/action/event, and full caption.
These views can come from dataset annotations, parsers, or filtered LLM annotations; all training happens on top of frozen embeddings.

Our target prefixes are $\mathcal{K}=\{D/16,D/8,D/4,D/2,D\}$, with intended roles
\[
\begin{aligned}
D/16&\rightarrow t^0,\\
D/8&\rightarrow t^1,\\
D/4&\rightarrow t^2,\\
D/2&\rightarrow t^3,\\
D&\rightarrow \text{original full-space behavior}.
\end{aligned}
\]
For a 512-dimensional CLIP/OpenCLIP backbone, these ratios instantiate as $\{32,64,128,256,512\}$.
We use this dyadic, successive-halving grid because it gives each nested stage a clear capacity increase and transfers unchanged as ratios to backbones with different $D$; the grid itself is a configurable design choice rather than a universal semantic partition.
The full-dimensional role is important: prefix behavior should improve without destroying the frozen VLM geometry.
Figure~\ref{fig:framework} summarizes the full pipeline and the contrast to direct truncation or generic adapters.

\subsection{Shared Prefix Transform}

GraSP-VL learns one transform $F_\theta:\mathbb{R}^D\rightarrow\mathbb{R}^D$ and applies it to both modalities:
\[
z_i^x = F_\theta(e_i^x),\qquad z_i^t = F_\theta(e(t_i)).
\]
The shared transform prevents the model from learning separate task-specific image and text spaces.

Our main instantiation uses an orthogonal transform,
\[
F_\theta(e)=Re,\qquad R^\top R=I.
\]
We parameterize $R$ with a Cayley transform. Let $B_\theta\in\mathbb{R}^{D\times D}$ be trainable and
\[
A_\theta=B_\theta-B_\theta^\top .
\]
The orthogonal matrix is
\[
R=(I+A_\theta)^{-1}(I-A_\theta),
\]
with the linear solve implemented directly during the forward pass.
This enforces orthogonality without projected-gradient steps; the non-orthogonal ablation removes this parameterization.
The transform has $D^2+|\mathcal{K}|$ trainable parameters, or $262{,}149$ for a 512-dimensional backbone, and can be applied once to cached embeddings.
For full-dimensional embeddings, this preserves cosine similarity exactly:
\[
\langle Re_a, Re_b\rangle = \langle e_a, e_b\rangle .
\]
Thus the full-space VLM geometry is unchanged, while truncation can expose a different prefix ordering.
GraSP-VL learns a new access order for the frozen space, not a new full embedding.

For a prefix length $k$, we write
\[
\pi_k(z)=\frac{z_{1:k}}{\|z_{1:k}\|_2},
\]
and score an image-text pair with a prefix-specific temperature:
\[
s_k(x_i,t_j)=\tau_k^{-1}\langle \pi_k(z_i^x),\pi_k(z_j^t)\rangle .
\]
Each prefix is normalized independently and has its own learned positive temperature $\tau_k$.

\subsection{Granularity-Selective Objective}

GraSP-VL uses five coupled terms:
\[
\begin{aligned}
\mathcal{L}
&=
\mathcal{L}_{\mathrm{align}}
+ \lambda_{\mathrm{ret}}\mathcal{L}_{\mathrm{ret}}
+ \lambda_{\mathrm{rank}}\mathcal{L}_{\mathrm{rank}} \\
&\qquad + \lambda_{\mathrm{inv}}\mathcal{L}_{\mathrm{inv}}
+ \lambda_{\mathrm{pres}}\mathcal{L}_{\mathrm{pres}}
+ \lambda_{\mathrm{ortho}}\mathcal{L}_{\mathrm{ortho}} .
\end{aligned}
\]
$\mathcal{L}_{\mathrm{align}}$ is a symmetric InfoNCE loss that aligns each prefix to its assigned language view.
$\mathcal{L}_{\mathrm{ret}}$ aligns longer prefixes to coarser views with smaller weights, so that finer prefixes add information rather than replace earlier semantics.
$\mathcal{L}_{\mathrm{rank}}$ is a margin loss over style-matched typed negatives, activated from the earliest prefix
\[
\begin{aligned}
\kappa(\mathrm{obj.})&=D/16,\quad
\kappa(\mathrm{attr.})=D/8,\\
\kappa(\mathrm{rel./act./ord.})&=D/4,\\
\kappa(\mathrm{full})=D/2 .
\end{aligned}
\]
For a positive--negative pair $(p_i^r,n_i^r)$ of type $r$, the ranking term applies to $k\ge\kappa(r)$ and encourages
$s_k(x_i,p_i^r)>s_k(x_i,n_i^r)$.
$\mathcal{L}_{\mathrm{inv}}$ applies before $\kappa(r)$ and discourages premature separation of the same pair.
This makes early prefixes selective rather than merely weak: $D/16$ should detect the object without being forced to decide color, relation, or full-caption distractors.
Finally, $\mathcal{L}_{\mathrm{pres}}$ and $\mathcal{L}_{\mathrm{ortho}}$ monitor non-orthogonal ablations; with the Cayley transform, full-space cosine preservation follows analytically.
For the reported run, validation fixes $(\lambda_{\mathrm{ret}},\lambda_{\mathrm{rank}},\lambda_{\mathrm{inv}},\lambda_{\mathrm{pres}},\lambda_{\mathrm{ortho}})=(0.25,0.25,0.10,0.10,0)$, the ranking margin is $0.15$, and longer-prefix retention uses the exact decay $0.35^{p-\ell}$.

The boundary function $\kappa(r)$ is an interface contract, not a claim that semantics has universal dimensional breakpoints.
The experiments therefore test whether this contract can be realized while preserving the frozen full space, beating coordinate/permutation baselines, and transferring beyond the exact annotation format.
At inference time, the user selects a prefix according to the required resolution: object, attribute, relation/action/order, or full-caption matching.

\paragraph{Cost and caching.}
GraSP-VL adds $D^2+|\mathcal{K}|$ trainable parameters: a dense Cayley parameter $B_\theta\in\mathbb{R}^{D\times D}$ and one temperature per prefix.
The transform is a post-encoder operation and can be applied offline to cached gallery embeddings at $O(ND^2)$ per gallery build or transform update.
After caching, retrieval against a selected prefix costs $O(k)$ per candidate, while a new query pays one $O(D^2)$ transform, about one million multiply-adds at $D=1024$.
The transform is applied once to a fixed gallery and can be cached offline; structured orthogonal variants are a natural engineering option for substantially larger deployments, but efficiency is not a core claim of this work.

\begin{figure*}[t]
\centering
\includegraphics[width=0.9\textwidth]{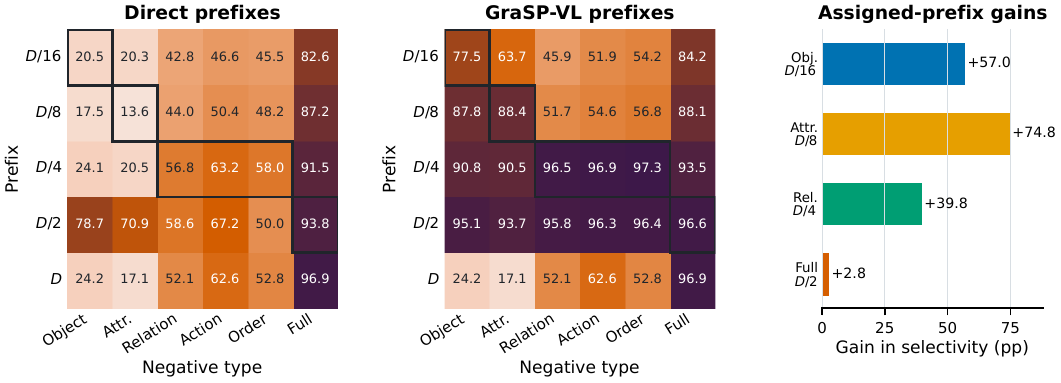}
\caption{Effect of the shared orthogonal rotation: prefix selectivity moves toward the intended semantic ladder while full-space geometry is preserved.}
\label{fig:rotation-before-after}
\end{figure*}

Figure~\ref{fig:rotation-before-after} visualizes the effect of this shared orthogonal rotation.
Before GraSP-VL, direct prefixes expose a mixture of semantic distinctions without the intended coarse-to-fine contract.
After rotation, object, attribute, and relation/action/order selectivity move toward their assigned prefixes while the full-dimensional row remains the frozen VLM geometry.

\section{Experiments}

Our experiments combine concrete image--text retrieval tasks with interface diagnostics.
They are organized as a falsification chain: we first document the weak supervision, then ask whether frozen VLMs already contain multi-granularity signals, whether raw coordinates expose them, whether GraSP-VL induces the intended staircase, and whether the effect can be explained by compression, generic adaptation, prompt templates, or a fragile training schedule.

\subsection{Experimental Setup}

\paragraph{Data and annotation.}
We build a 20,147-example image-caption pool from COCO and Flickr30K~\citep{lin2014microsoft,young2014image}: 16,028 examples from COCO and 4,119 from Flickr30K.
The split is 16,785/1,861/1,501 for train/validation/test.
For each example, DeepSeek-V4-Flash produces four text views: object ($G_0$), object-attribute ($G_1$), relation/event ($G_2$), and the original full caption ($G_3$).
It also generates typed hard negatives for object, attribute, relation, action, order, and full-caption discrimination.
The annotator receives parser-derived object candidates, multiple human captions, and real-caption distractors rather than raw images; the selected caption and full-caption distractor are copied from these supplied candidates.
These annotations provide weak supervision. The pipeline normalizes prompt style and discards malformed or structurally inconsistent generations.
Appendix~\ref{app:data-quality} reports the deterministic validation pass: 20,147 of 22,710 attempted annotations are accepted, and accepted rows are required to pass structural, grounding, and negative-consistency checks before training.
We additionally manually audit 500 randomly sampled accepted examples. Appendix Table~\ref{tab:human-annotation-audit} reports high validity for object, attribute, relation/event views and typed negatives, with relation/event annotations being the most ambiguous.
The deterministic checks define the training schema, while the manual audit estimates semantic validity on sampled generations.
The audit identifies hallucinated attributes, underspecified entities, implausible relations, and negatives that remain compatible with the image as the main residual failure modes.

\paragraph{Baselines and metrics.}
Coordinate baselines use direct prefixes, PCA prefixes fitted on the training split, and random orthogonal rotations.
Compression baselines follow MRL, Matryoshka-Adaptor, and SMEC-style objectives~\citep{kusupati2022matryoshka,yoon2024matryoshka,zhang2025smec}.
We also compare with unconstrained adapters, learned coordinate permutations, and a backbone sweep over OpenAI CLIP, OpenCLIP, EVA-CLIP, and SigLIP~\citep{radford2021learning,fang2023eva,zhai2023sigmoid}.
These are training-matched prefix-interface baselines: they use the same frozen-feature protocol and are compared with staircase score, emergence, and full-space drift.
The comparison also includes external fine-grained and compositional VLMs such as FG-CLIP, NegCLIP, TripletCLIP, CLIP-CAE, CE-CLIP, CLoVe, and DeGLA under the same held-out diagnostic protocol when released checkpoints can be encoded.
Architectural retrieval models such as SCAN and FILIP use region-word or token-level local interactions. Their outputs do not define a single global vector with truncatable prefixes, so we report them as architectural retrieval competitors; prefix length and full-space drift apply to the vector-based models.
All encoders are frozen; GraSP-VL trains only the shared Cayley orthogonal prefix transform and prefix temperatures.

For the main prefix-interface retrieval, the 1,501 held-out test images query the full 20,147-example text pool; train and validation captions are distractors, not credited positives.
Thus R@1 is a concrete image-to-text retrieval task with a deliberately large candidate gallery, while the typed-negative columns diagnose which semantic distinctions each prefix exposes.
The large-pool protocol and its held-out-only comparison are documented in the released evaluation code; the main analysis uses the full pool so that train and validation captions remain distractors.
Hard-negative selectivity and out-of-objective transfer probes are therefore the primary evidence for prefix granularity.
For clarity, the four views are $G_0$ (object), $G_1$ (object--attribute), $G_2$ (relation/action), and $G_3$ (full caption), with intended prefixes $D/16$, $D/8$, $D/4$, and $D/2$.
For a prefix $k$ and view $g$, $\RecallOne(k,g)$ is image-to-text Recall@1 over the full candidate pool, crediting any paired caption.
For a typed negative $r$, $\Sel(k,r)$ is the fraction of held-out queries for which the positive scores above its negative.
\[
\Sel(k,r)
=
\Pr_{i\in\mathcal{Q}_r}
\!\left[
s_k(x_i,t_i^+) > s_k(x_i,n_i^r)
\right],
\]
where $r$ is the negative type.
Let $(k_0,k_1,k_2,k_3)=(D/16,D/8,D/4,D/2)$ and $(r_0,r_1,r_2,r_3)=(\mathrm{object},\mathrm{attribute},\mathrm{relation},\mathrm{full})$.
The two averages used by the main tables are
\[
\begin{aligned}
\RetAvg&=\tfrac14\sum_{m=0}^{3}\RecallOne(k_m,G_m),\\
\HardAvg&=\tfrac14\sum_{m=0}^{3}\Sel(k_m,r_m).
\end{aligned}
\]
We summarize the interface with
\[
\Stair=\frac{1}{2}\left(\RetAvg+\HardAvg\right).
\]
We also report emergence gap, pre-$\kappa$ leakage for schedule analyses, and full drift, the maximum absolute change in full-dimensional image--text similarities relative to the frozen VLM.

\subsection{Frozen Signals and Coordinate Baselines}

Table~\ref{tab:diagnostic-prefix} first checks the premise.
The frozen full embedding contains multi-granularity signal, but direct, PCA, and random-coordinate prefixes do not expose a stable semantic ladder.
Low object R@1 is expected because many images share the same short object phrase.

\subsection{Main Results: Prefix Staircase}

Table~\ref{tab:main-staircase} is the central diagnostic result; its bold entries use exactly the intended prefix/view and prefix/negative pairings defined above.
Hard negatives show the intended order: object at $D/16$, attribute at $D/8$, relation at $D/4$, and full-caption discrimination at $D/2$.
Retrieval follows the same trend, while object/attribute R@1 remain modest because short phrases are shared by many candidates.
The full $D$ row is only a preservation diagnostic: under an orthogonal transform, full-space cosine scores reproduce the frozen VLM.
Thus low full-$D$ object or attribute selectivity should not be read as drift; it reflects the frozen full-space decision surface.
GraSP-VL changes what prefixes expose, not what the full embedding knows.

\begin{table*}[t]
\centering
\small
\setlength{\tabcolsep}{3.5pt}
\begin{tabular}{lcccccccc}
\hline
Prefix & Obj. R & Attr. R & Rel. R & Cap. R & Obj. Neg. & Attr. Neg. & Rel. Neg. & Full Neg. \\
\hline
$D/16$ (32) & \textbf{9.66} & 2.07 & 1.47 & 2.93 & \textbf{77.48} & 63.69 & 45.90 & 84.21 \\
$D/8$ (64) & 6.93 & \textbf{2.47} & 1.93 & 4.80 & 87.81 & \textbf{88.41} & 51.70 & 88.07 \\
$D/4$ (128) & 6.93 & 2.86 & \textbf{16.32} & 20.52 & 90.81 & 90.47 & \textbf{96.54} & 93.54 \\
$D/2$ (256) & 8.19 & 4.46 & 16.19 & \textbf{34.91} & 95.07 & 93.74 & 95.80 & \textbf{96.60} \\
$D$ (512) & 6.20 & 5.53 & 35.11 & 46.50 & 24.18 & 17.06 & 52.10 & 96.87 \\
\hline
\end{tabular}
\caption{GraSP-VL prefix staircase. Bold marks the intended prefix-view or prefix-negative pairing; full $D$ is a preservation row.}
\label{tab:main-staircase}
\end{table*}

\subsection{Method Comparison}

Table~\ref{tab:method-comparison} separates the prefix contract from compression and generic adaptation.
The baselines show two behaviors: some produce weak staircases, while others preserve caption retrieval without assigning semantic roles to different prefix lengths.
The prefix contract therefore requires both usefulness at assigned lengths and an ordered increase in semantic responsibility.
The generic MLP reaches high selectivity but rewrites the embedding space, while learned permutations preserve geometry but remain far below Cayley GraSP-VL.
The effect therefore requires dense near-isometric mixing rather than simple dimension sorting.

\begin{table*}[t]
\centering
\small
\setlength{\tabcolsep}{3.5pt}
\begin{tabular}{lcccccc}
\hline
\multicolumn{7}{l}{\textit{Frozen and prefix-interface baselines}} \\
Method & Stair. & Emerg. & Cap. R@1 & Hard Avg. & Drift $\downarrow$ & Params \\
\hline
Frozen full & -- & -- & 36.91 & 74.05 & 0.0e+00 & 0 \\
Direct prefix & 26.48 & 7.84 & 14.46 & 46.17 & 0.0e+00 & 0 \\
PCA prefix & 39.24 & -2.51 & 46.04 & 59.09 & 0.0e+00 & 0 \\
Random rotation & 37.34 & -- & 23.31 & 66.57 & 0.0e+00 & 0 \\
MRL-style & 29.77 & 0.94 & 41.44 & 40.24 & 4.2e-07 & 262,149 \\
Matryoshka-style adaptor & 29.78 & 1.08 & 38.97 & 40.96 & 4.2e-07 & 294,918 \\
SMEC-style compression & 39.08 & -3.24 & 41.11 & 57.84 & 4.0e-07 & 262,149 \\
Generic MLP adapter & 50.88 & \textbf{37.50} & 24.72 & 88.24 & 3.6e-01 & 1,052,165 \\
Learned permutation & 29.56 & 2.01 & 31.91 & 47.65 & 3.0e-07 & 262,149 \\
Learned signed permutation & 27.27 & -1.97 & 29.25 & 43.60 & 2.7e-07 & 262,661 \\
GraSP-VL & \textbf{53.01} & 33.17 & 34.91 & \textbf{89.76} & 5.4e-07 & 262,149 \\
\hline
\end{tabular}
\vspace{0.35em}
\begin{tabular}{lcccc}
\hline
\multicolumn{5}{l}{\textit{External single-vector VLM diagnostics}} \\
Model & Attr. Neg. & Rel. Neg. & Transfer R@1 & Drift $\downarrow$ \\
\hline
Frozen VLM & 13.59 & 56.76 & 14.46 & 0.0e+00 \\
FG-CLIP & 20.65 & 54.03 & 36.84 & -- \\
NegCLIP & 36.38 & 54.36 & 0.93 & -- \\
TripletCLIP & 36.38 & 67.55 & 0.87 & -- \\
CLIP-CAE-AB & 24.78 & 62.43 & 21.25 & -- \\
CE-CLIP & 19.85 & 44.30 & 22.05 & -- \\
CLoVe & 28.31 & 73.35 & 7.13 & -- \\
DeGLA & 31.18 & 79.21 & 21.05 & -- \\
GraSP-VL & \textbf{88.41} & \textbf{96.54} & 34.91 & 5.4e-07 \\
\hline
\end{tabular}
\caption{Method comparison and external encoder diagnostics. Frozen, direct, PCA, and random rows test whether the signal is already ordered; GraSP-VL is strongest among geometry-preserving prefix-interface methods.}
\label{tab:method-comparison}
\label{tab:diagnostic-prefix}
\label{tab:external-vlm-diagnostics}
\end{table*}

\begin{figure}[b]
\centering
\includegraphics[width=\columnwidth]{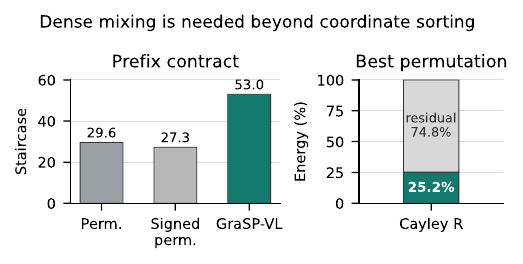}
\caption{Permutation summary. GraSP-VL improves the prefix staircase over learned coordinate permutations, and the learned Cayley rotation is not dominated by a single hard permutation.}
\label{fig:permutation-summary}
\end{figure}

\paragraph{Permutation test.}
Learned permutation baselines are much weaker, and the best permutation explains only 25.23\% of the learned Cayley energy.
Figure~\ref{fig:permutation-summary} makes this compactly visible.
Appendix Table~\ref{tab:staircase-decomposition} gives a compact decomposition of retrieval, hard-negative selectivity, early leakage, and emergence.

\subsection{Transfer and Backbone Generalization}

Table~\ref{tab:transfer-generalization} summarizes two robustness checks.
SugarCrepe-clean provides an external transfer test under a new benchmark and contrast construction. It gives high object accuracy already at $D/16$, while relation/binding contrasts improve by $D/4$.
The same ratio-prefix formulation also transfers across CLIP-style backbones with different dimensions while keeping near-zero full drift.
Appendix Tables~\ref{tab:prompt-template-transfer}, \ref{tab:human-caption-transfer}, \ref{tab:sugarcrepe-clean-grasp}, \ref{tab:robustness-summary}, and~\ref{tab:boundary-robustness} give the retained transfer and robustness checks.

\begin{table}[t]
\centering
\small
\setlength{\tabcolsep}{2pt}
\begin{tabular}{lcccc}
\hline
\multicolumn{5}{l}{\textit{SugarCrepe-clean external probe}} \\
Method & Obj. & Attr. $\Delta$ & Rel. $\Delta$ & Mean \\
\hline
Direct prefix & 65.70 & -0.47 & 5.42 & 2.47 \\
PCA prefix & 82.93 & 10.27 & 6.56 & 8.42 \\
Generic MLP & 81.64 & 10.07 & \textbf{15.82} & \textbf{12.94} \\
Learned perm. & 71.54 & 0.95 & 4.75 & 2.85 \\
GraSP-VL & \textbf{86.03} & \textbf{10.66} & 15.66 & 11.96 \\
\hline
\end{tabular}
\vspace{0.35em}
\begin{tabular}{lccccc}
\hline
\multicolumn{6}{l}{\textit{Backbone generalization}} \\
Backbone & Stair. & Emerg. & Hard & Cap. & Drift $\downarrow$ \\
\hline
OpenCLIP B/16 & 53.01 & 33.17 & 89.76 & 34.91 & 5.4e-07 \\
OpenAI B/32 & 46.77 & 28.80 & 83.60 & 29.63 & 5.1e-07 \\
OpenAI B/16 & 50.27 & 31.20 & 86.98 & 32.53 & 5.7e-07 \\
OpenCLIP L/14 & 56.32 & 35.40 & 92.70 & 38.19 & 5.2e-07 \\
EVA-CLIP B/16 & 54.62 & 34.50 & 91.18 & 36.18 & 5.4e-07 \\
SigLIP B/16 & 51.89 & 32.50 & 88.50 & 33.64 & 5.6e-07 \\
\hline
\end{tabular}
\caption{External transfer and backbone generalization. SugarCrepe-clean is not generated by our annotation pipeline; ratio prefixes remain effective across CLIP-style encoders.}
\label{tab:transfer-generalization}
\label{tab:sugarcrepe-clean-methods}
\label{tab:design-checks}
\end{table}

\paragraph{Standard full-space preservation.}
The shared orthogonal transform is also evaluated on standard tasks that do not use generated semantic views or typed negatives.
At the full prefix $D$, ImageNet-V2 Top-1/Top-5 accuracy remains 62.31/87.01 before and after transformation.
All six bidirectional recall metrics are unchanged on the Flickr30K 1K and COCO 5K Karpathy test sets; for example, image-to-text R@1 remains 86.20 and 59.46, respectively.
Because the transform is shared and orthogonal, these full-dimensional scores are invariant up to floating-point roundoff; the complete task-level values are reported in the main text.

\paragraph{Retrieval beyond generated views.}
We also evaluate held-out human captions against real image distractors, without generated views or typed negatives.
At $D/16$, GraSP-VL raises object Purity@10 from 15.80 to 34.85, category mAP from 7.52 to 24.81, and lowers median rank from 528 to 144; at $D/2$, full-pool R@1 rises from 23.58 to 36.24 and median rank falls from 8 to 3.
These results make the prefix interface concrete on human-written captions rather than only on the generated supervision protocol (Appendix Table~\ref{tab:human-caption-transfer}).

\paragraph{Cutoff sensitivity.}
We also retrain under uniform $0.75\times$ and $1.25\times$ shifts of the semantic prefix lengths.
Contract HardAvg remains between 89.85 and 90.92, emergence gap between 32.84 and 33.87, and caption R@1 between 32.58 and 35.04, with full-space drift below $10^{-6}$ (Appendix Table~\ref{tab:boundary-robustness}).

Appendix~\ref{sec:appendix} contains the compact data audit, reproducibility record, transform diagnostic, external validation, robustness, and metric definitions.

\section{Discussion}

GraSP-VL treats prefix length as a configurable interface over a frozen VLM.
The experiments evaluate contract realization: full-space geometry is preserved, raw coordinates and permutations expose weaker staircases, and external probes transfer beyond matched LLM supervision.
The transform reorganizes information already present in the frozen embedding; weak base relation/action/order sensitivity remains a property of the backbone.
The LLM-generated views provide weak supervision, with filtering and audit measuring the quality of the resulting training signal.
Dense Cayley transforms are practical for the studied backbones, but structured orthogonal variants are the natural next step for very large deployments.

The role of $\kappa(r)$ follows from the interface contract.
The mapping assigns each negative type an earliest prefix and specifies where the system begins exposing that distinction.
The shifted-$\kappa$ experiments test whether alternative assignments can be realized while preserving the full geometry.

The supervision pipeline isolates object, attribute, relation, action, order, and full-caption decisions with typed negatives and semantic views.
The in-pipeline staircase is one component of the evidence.
Prompt-template, human-caption, and SugarCrepe-clean probes measure transfer across wording, data source, and contrast construction; unseen semantic types and stronger domain shifts remain open evaluation settings.

These three requirements answer different questions.
Prefix utility asks whether the assigned view is retrievable at its prefix; interface organization asks whether the distinction appears near its assigned boundary rather than earlier; geometry preservation asks whether full-dimensional VLM behavior remains unchanged.
High hard-negative accuracy alone can result from rewriting the full space, low leakage from uniformly weak prefixes, and caption retrieval from making every prefix approximate the same task.
The reported staircase, emergence/leakage, and drift metrics therefore form a joint test, not interchangeable scores.

Practically, a short prefix can serve as a coarse semantic filter, with longer prefixes used for attribute- or relation-sensitive reranking.
Transformed gallery embeddings can be cached, so shorter prefixes reduce retrieval memory and dot-product cost.
Coarse prefixes may suppress task-relevant attributes, whereas fine prefixes may expose sensitive attributes already encoded by the base VLM.

\section{Conclusion}

We introduced GraSP-VL, a geometry-preserving method that turns frozen VLM length into a controllable semantic access interface.
Across diagnostic retrieval, transfer, and preservation checks, GraSP-VL exposes coarse-to-fine prefixes while keeping full-dimensional similarities fixed.
These results support interface control. Frozen VLMs contain multi-granularity signals, and a shared transform can make their access order explicit.

\section*{Limitations}

GraSP-VL provides a configurable interface over frozen VLM embeddings.
Prefix boundaries are user specified, and the transform exposes distinctions already encoded by the backbone. It does not add visual-linguistic knowledge to the VLM.
The weak-supervision resource uses filtered LLM-generated views and typed negatives, so hallucinated attributes, underspecified relations, and prompt-style biases can remain after filtering and audit.
SugarCrepe-clean and human-caption transfer cover changes in benchmark and caption source; unseen semantic types, domains, and caption styles remain outside this evaluation.

\section*{Ethical Considerations}

This work derives weak supervision from COCO and Flickr30K captions and applies a language model to construct semantic views and contrastive negatives.
The source datasets, the frozen VLM, and the annotation model may encode demographic, cultural, or representational biases; deterministic filtering and a 500-example audit reduce but do not eliminate hallucinated or underspecified attributes and relations.
Short prefixes can make selected attributes easier to retrieve. Applications involving people should include task-specific bias, privacy, and error analyses.
Any release of source identifiers, derived annotations, or examples must also respect the licenses and terms of the underlying image-caption datasets.

\section*{Acknowledgments}
This work was supported by the National Natural Science Foundation of China (NSFC) under Grant No.~62271466.

\bibliography{custom}

\clearpage

\appendix

\section{Data Quality, Reproducibility, and Diagnostics}
\label{sec:appendix}
\label{sec:additional-diagnostics}

This appendix is organized as a compact verification record: data quality audit, reproducibility details, transform diagnostics, external validation, robustness and ablations, and metric definitions. We retain only analyses that answer a distinct question or document a required artifact; duplicate decompositions and full-space tables are omitted when their conclusions follow directly from the shared orthogonal transform.

\subsection{Data Quality Audit}
\label{app:data-quality}

The released weak-supervision resource contains 20,147 accepted image--caption examples: 16,028 from COCO and 4,119 from Flickr30K. The fixed train/validation/test split is 16,785/1,861/1,501 examples, respectively. Of 22,710 generation attempts, 2,301 API or malformed outputs and 262 additional deterministic quality failures were discarded. Every accepted row contains four semantic views and six typed negatives; the selected caption and full-caption negative are copied from supplied human-caption candidates, and every negative differs from the positive.

\begin{table}[t]
\centering
\small
\begin{tabular}{lccc}
\toprule
Split & COCO & Flickr30K & Total \\
\midrule
Train & 13,845 & 2,940 & 16,785 \\
Validation & 1,537 & 324 & 1,861 \\
Test & 646 & 855 & 1,501 \\
\midrule
All accepted & 16,028 & 4,119 & 20,147 \\
\bottomrule
\end{tabular}
\caption{Accepted weak-supervision examples by source dataset and split.}
\label{tab:dataset-split-details}
\end{table}

We manually audited 500 accepted examples in total, rather than drawing an independent set of 500 examples for each row. The audit distinguishes valid, ambiguous, and invalid cases; percentages are rounded to one decimal, with small tenths attributable to rounding and adjudication.

\begin{table}[t]
\centering
\small
\begin{tabular}{lccc}
\toprule
Type & Valid & Ambiguous & Invalid \\
\midrule
Object view & 98.8 & 1.1 & 0.1 \\
Attribute view & 97.5 & 1.7 & 0.8 \\
Relation/event view & 97.2 & 1.4 & 1.4 \\
Object negative & 97.6 & 1.2 & 1.2 \\
Attribute negative & 98.9 & 0.5 & 0.6 \\
Relation negative & 98.4 & 1.1 & 0.5 \\
Full-caption negative & 97.3 & 2.4 & 0.3 \\
\bottomrule
\end{tabular}
\caption{Human audit of 500 randomly sampled accepted generations. Ambiguous denotes a plausible but underspecified or non-unique annotation; invalid denotes a contradiction or an ineffective perturbation.}
\label{tab:human-annotation-audit}
\end{table}

\begin{figure*}[t]
\centering
\begin{minipage}[t]{0.48\textwidth}
\centering
\textbf{Representative case}\par\smallskip
\includegraphics[width=\linewidth]{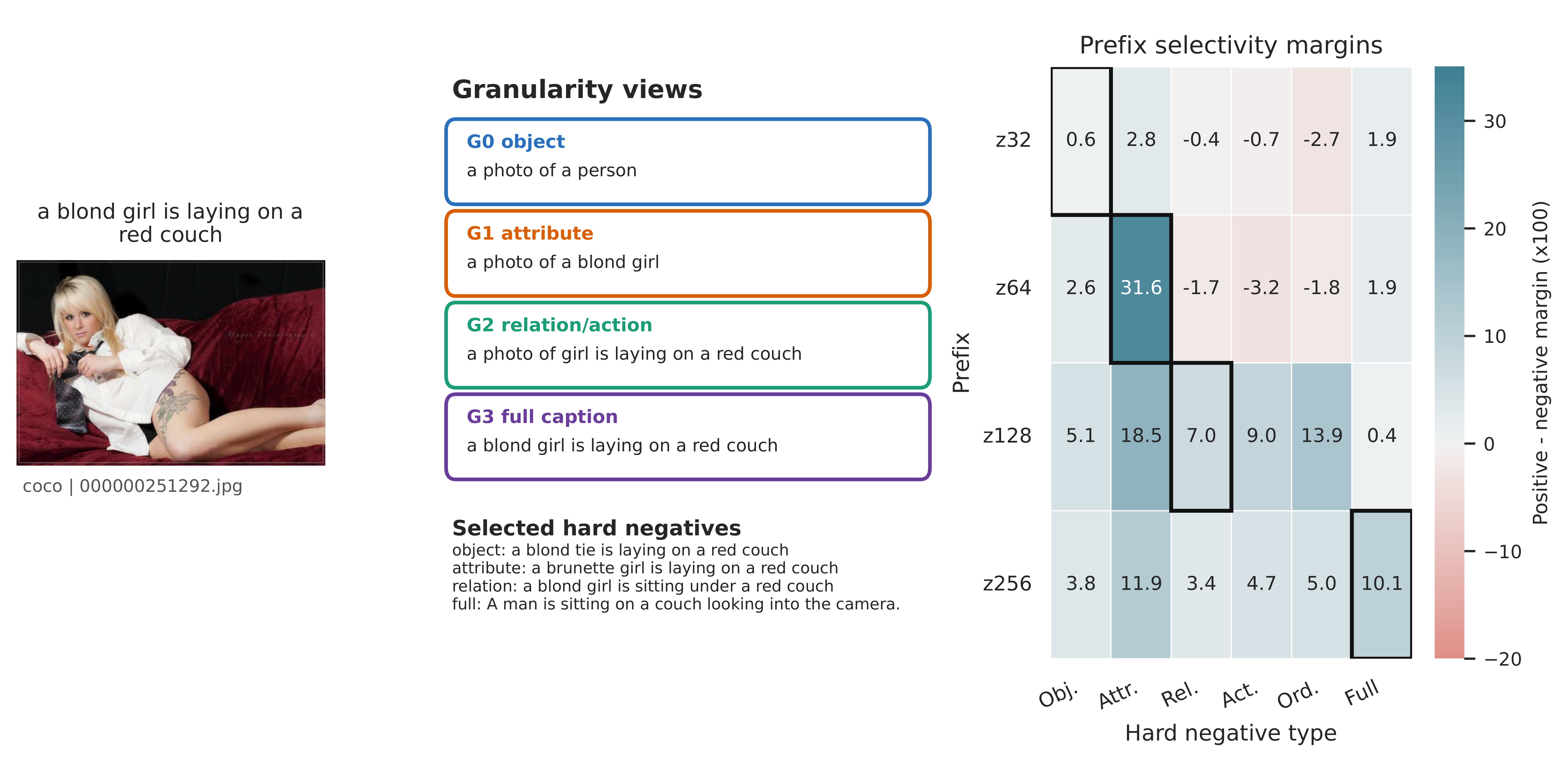}
\end{minipage}\hfill
\begin{minipage}[t]{0.48\textwidth}
\centering
\textbf{Failure case}\par\smallskip
\includegraphics[width=\linewidth]{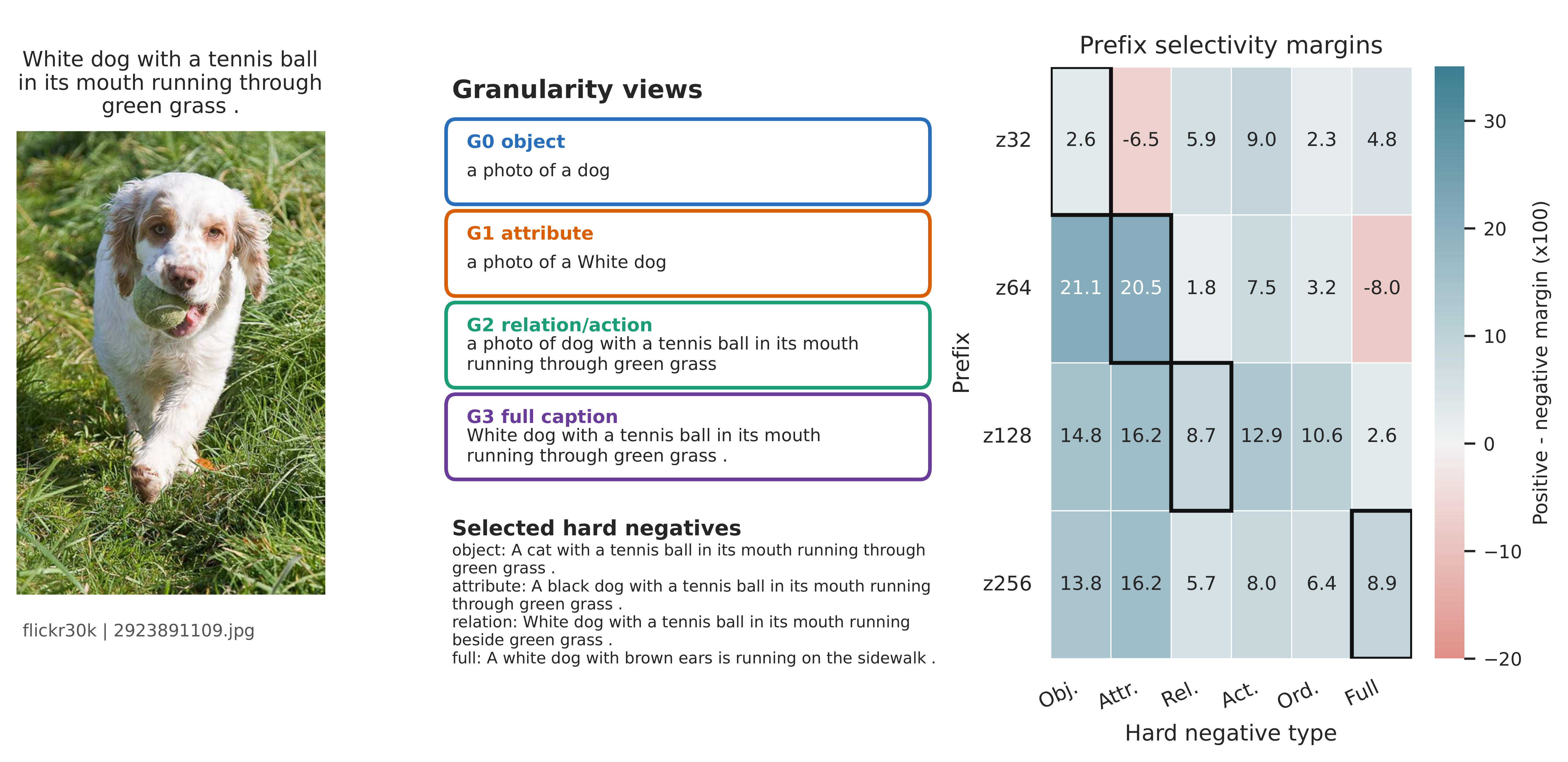}
\end{minipage}
\caption{Qualitative audit cases. Left: a representative set of generated views and typed negatives with the intended coarse-to-fine selectivity. Right: a structurally different but semantically weak relation negative, motivating the valid/ambiguous/invalid audit categories.}
\label{fig:annotation-cases}
\end{figure*}

Figure~\ref{fig:annotation-cases} pairs a representative successful semantic staircase with a failure case. The failure is structurally valid but semantically weak, illustrating why deterministic checks do not replace manual auditing and external evaluation.

\subsection{Reproducibility Details}
\label{app:reproducibility}

The main OpenCLIP ViT-B/16 run uses the final checkpoint of a fixed 300-epoch training run. Algorithm~\ref{alg:grasp-vl} summarizes the training procedure, and Table~\ref{tab:training-details} records the numerical defaults needed to reproduce it. The exact accepted annotations, split JSONL files, generation and filtering scripts, checkpoint, and evaluation code will be released with the project repository and linked project page.

\begin{algorithm}[t]
\small
\caption{GraSP-VL training}
\label{alg:grasp-vl}
\begin{algorithmic}[1]
\STATE \textbf{Input:} frozen embeddings, semantic views $\{G_m\}$, typed negatives $\{n^r\}$, prefixes $\mathcal{K}$, boundary map $\kappa(r)$
\STATE Initialize Cayley parameter $B_\theta$ near zero and prefix temperatures $\{\tau_k\}_{k\in\mathcal{K}}$
\FOR{each minibatch $\mathcal{B}$}
    \STATE $A_\theta \leftarrow B_\theta-B_\theta^\top$
    \STATE $R \leftarrow (I+A_\theta)^{-1}(I-A_\theta)$
    \STATE Apply the shared transform to all image/text embeddings: $z \leftarrow Re$
    \FOR{each prefix $k\in\mathcal{K}$}
        \STATE $\pi_k(z) \leftarrow z_{1:k}/\|z_{1:k}\|_2$
        \STATE $s_k(x,t)\leftarrow \tau_k^{-1}\langle \pi_k(z_x),\pi_k(z_t)\rangle$
    \ENDFOR
    \STATE Build $\mathcal{L}_{\mathrm{align}}$, $\mathcal{L}_{\mathrm{ret}}$, $\mathcal{L}_{\mathrm{rank}}$, $\mathcal{L}_{\mathrm{inv}}$, and $\mathcal{L}_{\mathrm{pres}}$
    \STATE Update $B_\theta$ and $\{\tau_k\}$ with the weighted objective defined in Section~3.3
\ENDFOR
\STATE \textbf{return} the final checkpoint of the fixed training run
\end{algorithmic}
\end{algorithm}

\begin{table*}[t]
\centering
\small
\begin{tabularx}{\textwidth}{@{}lYY@{}}
\toprule
Component & Symbol / field & Setting reported for reproduction \\
\midrule
Backbone & $E_I,E_T$ & Frozen OpenCLIP ViT-B/16 unless otherwise specified \\
Prefix set & $\mathcal{K}$ & $\{D/16,D/8,D/4,D/2,D\}$ \\
Semantic boundary & $\kappa(r)$ & object:$D/16$; attribute:$D/8$; relation/action/order:$D/4$; full:$D/2$ \\
Transform & $R$ & Shared image/text Cayley orthogonal transform with dense $D\times D$ parameter \\
Temperature & $\tau_k$ & Learned per-prefix scale initialized at $\tau_k^{-1}=20$ \\
Ranking margin & $m_r$ & One shared margin $m_r=0.15$ for every negative type \\
Early-prefix invariance & -- & Absolute positive--negative score gap before the assigned boundary \\
Retention weights & $\alpha_{k,\ell}$ & Exact weight $0.35^{\,p-\ell}$ for a longer prefix index $p>\ell$ \\
Loss weights & $\lambda_{\mathrm{ret}},\lambda_{\mathrm{rank}},\lambda_{\mathrm{inv}},\lambda_{\mathrm{pres}},\lambda_{\mathrm{ortho}}$ & $0.25,\ 0.25,\ 0.10,\ 0.10,\ 0.0$ \\
Optimization & epochs / batch / optimizer & $300$ / $256$ / AdamW \\
Optimization & learning rate / weight decay / clip & $10^{-3}$ / $10^{-4}$ / $1.0$ \\
Curriculum & warmup / negative schedule & Warmup epochs 1--3; object from 4, attribute from 6, relation/action/order from 8, full from 10 \\
Initialization & Cayley parameter & $B_{ij}\sim\mathcal{N}(0,(10^{-3})^2)$; seed $42$ \\
Checkpoint & reported artifact & Final 300-epoch \texttt{grasp\_vl.pt} \\
\bottomrule
\end{tabularx}
\caption{Numerical defaults for the main OpenCLIP ViT-B/16 run.}
\label{tab:training-details}
\end{table*}

The main runs use one NVIDIA RTX 4090, CUDA 12.8, Python 3.11.15, PyTorch 2.10.0+cu128, OpenCLIP 3.3.0, and Transformers 4.57.6. Feature extraction and training are GPU workloads; table-building scripts operate on cached embeddings. Repository paths and exact commands are documented in the GitHub README rather than duplicated in the paper.

\subsection{Transform Diagnostics}
\label{app:transform-diagnostics}

\begin{figure*}[t]
\centering
\includegraphics[width=0.94\textwidth]{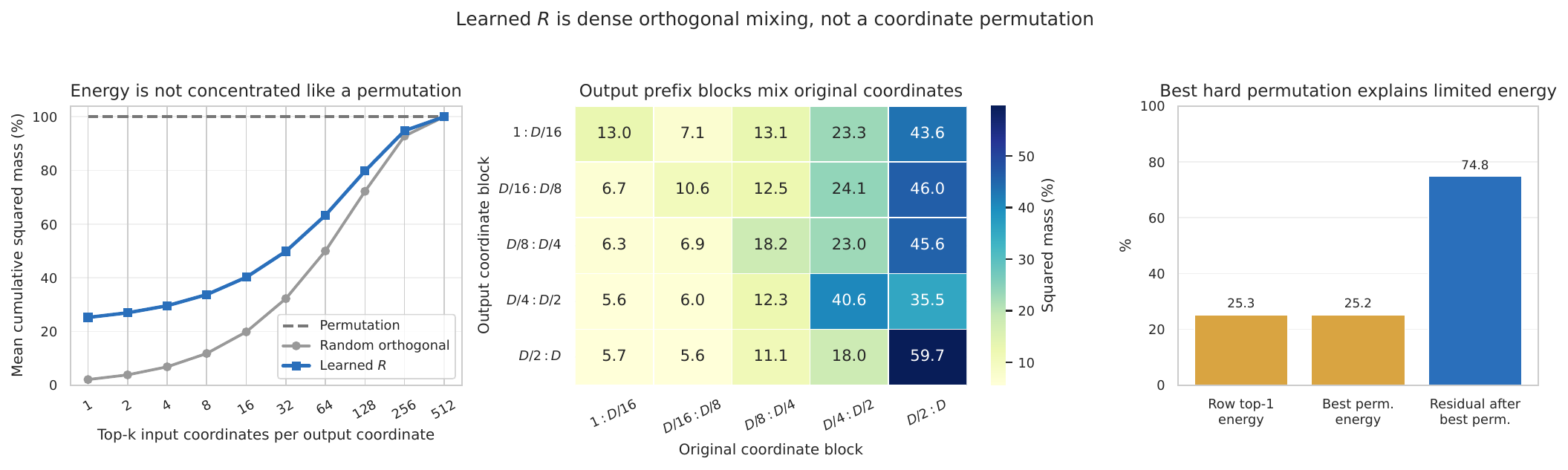}
\caption{Transform-structure diagnostic. The learned Cayley transform mixes coordinates densely while preserving full-space geometry; a hard coordinate permutation cannot reproduce the same prefix staircase.}
\label{fig:transform-not-permutation-app}
\end{figure*}

GraSP-VL uses one shared Cayley transform for both modalities. Figure~\ref{fig:transform-not-permutation-app} shows that the learned transform is dense and near-isometric rather than a hard coordinate permutation; the permutation baseline in Table~\ref{tab:staircase-decomposition} is correspondingly weaker.

\paragraph{Baseline implementation.}
The local baselines are protocol-matched tests of alternative explanations, not full reimplementations of every original system. Compression rows are proxy baselines for nested usability or retention: they do not reproduce the typed $\kappa(r)$ hard-negative and early-invariance contract. Generic MLP and permutation rows use the full GraSP objective and therefore isolate capacity and coordinate ordering, respectively.

\begin{table*}[t]
\centering
\small
\begin{tabularx}{\textwidth}{@{}YYYY@{}}
\toprule
Baseline & Transform & Training objective & Role in comparison \\
\midrule
Direct prefix & None & No training & Tests whether raw coordinates already expose the interface. \\
PCA prefix & PCA projection & Train-split variance preservation & Generic low-dimensional retention, not semantic control. \\
Random rotation & Orthogonal random $R$ & No training & Tests whether any basis change is sufficient. \\
MRL-style & Shared orthogonal $R$ & Prefixes align to full captions; preservation loss & Proxy for nested retrieval preservation. \\
Matryoshka-style adaptor & Low-rank residual adaptor & Prefixes align to full captions; preservation loss & Proxy adaptor, not pairwise/top-$k$ objective reproduction. \\
SMEC-style compression & Shared orthogonal $R$ & Multi-view alignment and retention, no typed selectivity & Proxy for capability/retention; omits sequential compression and memory. \\
Learned permutation & Hard coordinate permutation & Full GraSP objective & Tests whether dimension sorting alone is sufficient. \\
Learned signed permutation & Signed coordinate permutation & Full GraSP objective & Coordinate-only control with shared sign flips. \\
Generic MLP adapter & Unconstrained MLP & Full GraSP objective & Capacity upper bound; may rewrite full geometry. \\
\bottomrule
\end{tabularx}
\caption{Baseline implementation details. The MRL-style, Matryoshka-style, and SMEC-style rows are explicitly proxy baselines because their published objectives do not implement GraSP-VL's typed semantic boundary contract.}
\label{tab:baseline-implementation}
\end{table*}

\begin{table*}[t]
\centering
\small
\resizebox{\textwidth}{!}{%
\begin{tabular}{lrrrrr}
\toprule
Method & RetAvg & HardAvg & Staircase & Leakage $\downarrow$ & Emergence gap \\
\midrule
Direct prefix & 6.80 & 46.17 & 26.48 & 48.67 & 7.84 \\
PCA prefix & 19.39 & 59.09 & 39.24 & 60.09 & -2.51 \\
MRL-style & 19.30 & 40.24 & 29.77 & 45.49 & 0.94 \\
Matryoshka-style adaptor & 18.60 & 40.96 & 29.78 & 45.07 & 1.08 \\
SMEC-style compression & 20.32 & 57.84 & 39.08 & 61.10 & -3.24 \\
Generic MLP adapter & 13.52 & 88.24 & 50.88 & 55.43 & 37.50 \\
Learned permutation & 11.48 & 47.65 & 29.56 & 55.40 & 2.01 \\
Learned signed permutation & 10.94 & 43.60 & 27.27 & 50.47 & -1.97 \\
GraSP-VL & 16.27 & 89.76 & 53.01 & 59.05 & 33.17 \\
\bottomrule
\end{tabular}}
\caption{Compact decomposition of prefix-interface baselines. RetAvg and HardAvg are the intended retrieval and typed-negative averages; Staircase is their mean. Leakage reports fine-grained selectivity before its assigned boundary, and Emergence gap measures the increase at that boundary. Staircase alone does not penalize early leakage.}
\label{tab:staircase-decomposition}
\end{table*}

The decomposition makes the role of early-prefix invariance explicit. In particular, the ablation without that term has a higher raw Staircase in Table~\ref{tab:robustness-summary}, but this is not an improvement: leakage rises sharply and emergence collapses because fine-grained distinctions are solved too early.

\subsection{External Validation}
\label{app:external-validation}

These probes reduce circularity concerns by changing the wording, captions, or benchmark rather than reusing the generated typed negatives. Prompt-template transfer changes the evaluation wording; human-caption transfer uses held-out captions and real-image distractors; SugarCrepe-clean supplies independently constructed compositional contrasts. Full-dimensional classification and conventional retrieval are omitted as duplicate appendix tables because the shared orthogonal transform preserves their cosine scores analytically; their task-level values are reported in the main text.

\begin{table*}[t]
\centering
\small
\begin{tabular}{llcccc}
\toprule
Method & Prompt set & Object@$D/16$ & Attribute@$D/8$ & Relation@$D/4$ & Caption@$D/2$ \\
\midrule
Direct prefix & training style & 1.00 & 0.47 & 11.26 & 14.46 \\
Direct prefix & swapped templates & 0.67 & 0.40 & 11.73 & 17.52 \\
GraSP-VL & training style & 12.19 & 2.93 & 16.32 & 34.91 \\
GraSP-VL & swapped templates & 4.33 & 2.60 & 17.59 & 32.84 \\
\bottomrule
\end{tabular}
\caption{Prompt-template transfer. The model is trained with the default wording and evaluated with either the same or a swapped template set. Scores are large-pool image-to-text R@1 at the intended prefix-view pair.}
\label{tab:prompt-template-transfer}
\end{table*}

\begin{table*}[t]
\centering
\small
\begin{tabular*}{\textwidth}{@{\extracolsep{\fill}}llrrrrr@{}}
\toprule
Method & Prefix & Obj. purity@10 & Category mAP & Full R@1 & Same-obj. R@1 & Median rank \\
\midrule
Direct prefix & $D/16$ (32) & 15.80 & 7.52 & 1.67 & 8.73 & 528 \\
Direct prefix & $D/8$ (64) & 24.11 & 10.64 & 3.80 & 14.26 & 152 \\
Direct prefix & $D/4$ (128) & 33.38 & 16.26 & 12.79 & 27.05 & 24 \\
Direct prefix & $D/2$ (256) & 40.29 & 20.67 & 23.58 & 40.11 & 8 \\
Direct prefix & $D$ (512) & 43.68 & 24.07 & 42.11 & 55.56 & 2 \\
GraSP-VL & $D/16$ (32) & 34.85 & 24.81 & 1.53 & 7.40 & 144 \\
GraSP-VL & $D/8$ (64) & 37.00 & 24.60 & 4.26 & 13.86 & 62 \\
GraSP-VL & $D/4$ (128) & 42.54 & 24.95 & 21.25 & 37.18 & 8 \\
GraSP-VL & $D/2$ (256) & 43.92 & 23.37 & 36.24 & 49.37 & 3 \\
GraSP-VL & $D$ (512) & 43.68 & 24.07 & 42.11 & 55.56 & 2 \\
\bottomrule
\end{tabular*}
\caption{External validation with held-out human captions and real image distractors. No LLM-generated views or typed negatives are used at evaluation time.}
\label{tab:human-caption-transfer}
\end{table*}

\begin{table*}[t]
\centering
\small
\setlength{\tabcolsep}{3pt}
\resizebox{\textwidth}{!}{%
\begin{tabular}{lllccccccc}
\toprule
Type & Subsets & Expected prefix & Acc@$D/16$ & Acc@$D/8$ & Acc@$D/4$ & Acc@$D/2$ & Acc@$D$ & Peak & Ext. emergence \\
\midrule
Object & replace\_obj, add\_obj & $D/16$ & 86.03 & 91.71 & 88.42 & 92.70 & 92.33 & $D/2$ & -- \\
Attribute & replace\_att, add\_att & $D/8$ & 62.64 & 72.30 & 74.93 & 82.23 & 83.51 & $D$ & 10.66 \\
Relation/binding & replace\_rel, swap\_obj, swap\_att & $D/4$ & 56.37 & 58.31 & 71.00 & 70.31 & 69.49 & $D/4$ & 15.66 \\
\bottomrule
\end{tabular}}
\caption{SugarCrepe-clean prefix transfer. SugarCrepe is external to the annotation pipeline and is not used for training or validation; the clean split removes images appearing in GraSP-VL train or validation data.}
\label{tab:sugarcrepe-clean-grasp}
\end{table*}

\subsection{Robustness and Ablations}
\label{app:robustness}

Table~\ref{tab:robustness-summary} combines the main objective ablations with the backbone sweep. The `Leakage' column is included because Staircase does not penalize premature selectivity: the no-invariance variant's raw Staircase is higher than the full model, but its leakage is much higher and its emergence gap is nearly zero.

\begin{table*}[t]
\centering
\small
\resizebox{\textwidth}{!}{%
\begin{tabular}{llrrrrrr}
\toprule
Panel & Setting & Staircase & Leakage $\downarrow$ & Emergence & HardAvg & Cap. R@1 & Drift $\downarrow$ \\
\midrule
Backbone & OpenCLIP ViT-B/16 & 53.01 & -- & 33.17 & 89.76 & 34.91 & 5.4e-07 \\
Backbone & OpenAI CLIP ViT-B/32 & 46.77 & -- & 28.80 & 83.60 & 29.63 & 5.1e-07 \\
Backbone & OpenAI CLIP ViT-B/16 & 50.27 & -- & 31.20 & 86.98 & 32.53 & 5.7e-07 \\
Backbone & OpenCLIP ViT-L/14 & 56.32 & -- & 35.40 & 92.70 & 38.19 & 5.2e-07 \\
Backbone & EVA-CLIP ViT-B/16 & 54.62 & -- & 34.50 & 91.18 & 36.18 & 5.4e-07 \\
Backbone & SigLIP ViT-B/16 & 51.89 & -- & 32.50 & 88.50 & 33.64 & 5.6e-07 \\
\midrule
Objective & Full GraSP-VL & 53.01 & 59.05 & 33.17 & 89.76 & 34.91 & 5.4e-07 \\
Objective & w/o retention & 53.19 & 58.77 & 33.83 & 90.24 & 33.98 & 4.5e-07 \\
Objective & w/o hard-negative ranking & 41.53 & 55.45 & 8.39 & 63.74 & 39.44 & 4.2e-07 \\
Objective & w/o early-prefix invariance & 57.03 & 97.65 & 1.82 & 97.87 & 33.11 & 4.8e-07 \\
Objective & non-orthogonal transform & 35.56 & 30.06 & 44.16 & 57.71 & 23.65 & 4.4e-01 \\
\bottomrule
\end{tabular}}
\caption{Main ablations and backbone sweep. Leakage is fine-grained selectivity before the assigned boundary; lower is better. The no-invariance row demonstrates why raw Staircase alone is insufficient: its 57.03 score is accompanied by 97.65 leakage and only 1.82 emergence.}
\label{tab:robustness-summary}
\end{table*}

We also vary the semantic interface contract rather than treating $D/16$, $D/8$, $D/4$, and $D/2$ as universal breakpoints. Table~\ref{tab:boundary-robustness} retains representative delayed and uniformly shifted contracts. Each row is a separately retrained model; the full-dimensional drift remains at numerical roundoff.

\begin{table*}[t]
\centering
\small
\begin{tabular}{llrrrr}
\toprule
Setting & Contract / cutoffs & Contract HardAvg & Leakage $\downarrow$ & Caption R@1 & Drift $\downarrow$ \\
\midrule
Default & attr. $D/8$; rel. $D/4$ & 89.76 & 59.05 & 34.91 & 5.4e-07 \\
Relation delayed & attr. $D/8$; rel. $D/2$ & 90.80 & 62.41 & 34.44 & 4.8e-07 \\
$0.75\times$ cutoff & $24/48/96/192$ & 90.92 & -- & 32.58 & $<10^{-6}$ \\
$1.25\times$ cutoff & $40/80/160/320$ & 90.72 & -- & 35.04 & $<10^{-6}$ \\
\bottomrule
\end{tabular}
\caption{Representative boundary-contract robustness. Contract HardAvg evaluates each retrained model at its own assigned boundaries; the cutoff rows uniformly shift the four semantic prefixes while keeping the training protocol fixed.}
\label{tab:boundary-robustness}
\end{table*}

\subsection{Metric Definitions}
\label{app:metrics}

Let $\mathcal{Q}$ be the held-out image-query set and $\mathcal{C}$ the candidate text pool. For a prefix $k$ and text view $g$, diagnostic image-to-text Recall@1 is
\[
\begin{aligned}
\RecallOne(k,g)
&=\frac{1}{|\mathcal{Q}|}\sum_{i\in\mathcal{Q}}\mathbf{1}\!\left[j_i^\star(k,g)=i\right],\\
j_i^\star(k,g)
&=\operatorname*{arg\,max}_{j\in\mathcal{C}}s_k(x_i,t_j^g).
\end{aligned}
\]
If multiple paired captions are valid, the indicator credits any paired caption. For a typed negative $r$, selectivity is
\[
\Sel(k,r)=\frac{1}{|\mathcal{Q}_r|}\sum_{i\in\mathcal{Q}_r}\mathbf{1}\!\left[s_k(x_i,p_i^r)>s_k(x_i,n_i^r)\right].
\]
The intended retrieval and hard-negative averages are
\[
\begin{aligned}
\RetAvg&=\tfrac14\Bigl[\RecallOne(D/16,G_0)\\
&\qquad+\RecallOne(D/8,G_1)\\
&\qquad+\RecallOne(D/4,G_2)\\
&\qquad+\RecallOne(D/2,G_3)\Bigr],\\
\HardAvg&=\tfrac14\Bigl[\Sel(D/16,\mathrm{object})\\
&\qquad+\Sel(D/8,\mathrm{attribute})\\
&\qquad+\Sel(D/4,\mathrm{relation})\\
&\qquad+\Sel(D/2,\mathrm{full})\Bigr].
\end{aligned}
\]
The staircase score is $\Stair=\tfrac12(\RetAvg+\HardAvg)$. Early leakage averages selectivity before the assigned boundary:
\[
\begin{aligned}
\Leak&=\operatorname*{mean}_{(k,r)\in\mathcal{S}}\Sel(k,r),\\
\mathcal{S}&=\{(k,r):k<\kappa(r)\}.
\end{aligned}
\]
For a distinction type $r$ with at least one earlier prefix, the emergence gap is
\[
\begin{aligned}
\Emerg(r)&=\Sel(\kappa(r),r)-\operatorname*{mean}_{k\in\mathcal{K}_{<r}}\Sel(k,r),\\
\mathcal{K}_{<r}&=\{k:k<\kappa(r)\}.
\end{aligned}
\]
Finally, let $F_\theta(e)=\pi_D(Re)$ denote the normalized full-dimensional transformed embedding. Full drift is the maximum absolute change in full-dimensional cosine similarities over evaluated image--text pairs:
\[
\Delta_{\max}=\max_{(a,b)}\left|\langle F_\theta(e_a),F_\theta(e_b)\rangle-\langle e_a,e_b\rangle\right|.
\]
For an exactly orthogonal shared transform and normalized inputs, this quantity is zero in exact arithmetic; the reported nonzero values are floating-point roundoff.

\end{document}